%% file: main.tex
\pdfoutput=1

\documentclass[10pt,twocolumn,letterpaper]{article}

\PassOptionsToPackage{table}{xcolor}

\usepackage[pagenumbers]{cvpr}   % arXiv / preprint version

\usepackage{stfloats}
\usepackage{cuted}
\usepackage{caption}
\usepackage{booktabs}
\usepackage{multirow}


\definecolor{cvprblue}{rgb}{0.21,0.49,0.74}
\usepackage[
    breaklinks,
    colorlinks,
    allcolors=cvprblue
]{hyperref}

\def\confName{CVPR}
\def\confYear{2027}

\title{
DetailScale-SR: Decoupling Restoration and Continuous Detail Control
for Real-World Super-Resolution
}

\author{
Shengchuan Gao$^{1,2}$ \quad
Zhongjie Duan$^{2}$ \quad
Yingda Chen$^{2}$ \quad
Jingjing Chen$^{3}$\\[3pt]
$^{1}$Shanghai Jiao Tong University \quad
$^{2}$ModelScope Team, Alibaba Group \quad
$^{3}$Fudan University\\[3pt]
{\tt\small
gscdy111@sjtu.edu.cn \quad
\{duanzhongjie.dzj,yingda.chen\}@alibaba-inc.com}\\
{\tt\small chenjingjing@fudan.edu.cn}
}

\begin{document}

\maketitle

\begin{strip}
  \centering
  \includegraphics[width=\textwidth]{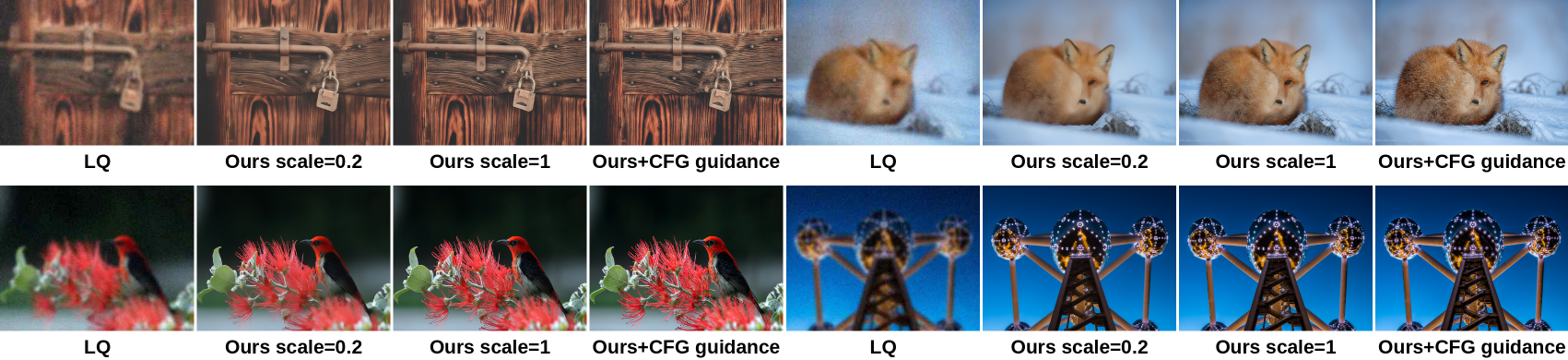}
  \captionof{figure}{
  Qualitative results of DetailScale-SR.
  Besides recovering fine textures and structures from heavily degraded
  inputs, our method enables detail-controllable SR through a single scalar
  that denotes the desired richness of reconstructed detail, while
  Detail-CFG extrapolates the learned direction beyond the ground truth.
  More results are shown in the supplementary material.
  }
  \label{fig:teaser}
\end{strip}

\input{sec/0_abstract}
\input{sec/1_intro}
\input{sec/2_relatedwork}
\input{sec/3_method}
\input{sec/4_Experiments}
\input{sec/5_conclusion}

\section*{LLM Usage Statement}

OpenAI ChatGPT and Codex were used to assist with language editing,
grammatical refinement, LaTeX formatting, bibliography formatting, citation
placement, and the layout and labeling refinement of figures. They were not
used to generate experimental data or numerical results. All references,
scientific claims, equations, analyses, and final manuscript content were
independently verified and approved by the authors, who take full
responsibility for the submission.

{
    \small
    \bibliographystyle{ieeenat_fullname}
    \bibliography{main}
}

% Optional supplementary material:
% \clearpage
% \appendix
% \input{sec/X_suppl}

\end{document}

%% file: sec/0_abstract.tex
\begin{abstract}
Real-world image super-resolution often seeks the most detailed plausible reconstruction, although more detail is not always desirable. The same degraded image may require restrained restoration, faithful reconstruction, or enhanced visual detail according to user intent. We present DetailScale-SR, a preference-conditioned framework that continuously controls reconstructed detail for a fixed low-resolution input without altering the scene being restored. Rather than relying on post-processing or repeatedly rewriting prompts, it learns a continuous family of reconstructions while separating content recovery from detail preference. This enables a smooth transition from conservative restoration to richly textured results under consistent input and semantic constraints. Its controllable range also extends beyond conventional restoration: DetailScale-SR can enhance high-frequency structures and generate perceptually richer textures than the paired high-resolution reference. Text restoration is another distinct strength. For text-rich scenes, especially Chinese posters and signs, it recovers more legible and character-faithful text than prior generative SR methods under matched textual guidance. DetailScale-SR therefore unifies continuous detail control, beyond-reference enhancement, and reliable recovery of semantically important content. Our model is available at
\url{https://modelscope.cn/models/DiffSynth-Studio/Qwen-Image-2.1-SR}.
\end{abstract}

%% file: sec/1_intro.tex
\section{Introduction}
\label{sec:intro}

Real-world image super-resolution aims to reconstruct a high-resolution image from a low-resolution observation affected by unknown combinations of blur, noise, compression, and downsampling. Since these degradations inevitably remove part of the high-frequency information, the desired fine-scale structures cannot always be determined from the observed pixels alone. Recent diffusion- and flow-based methods~\cite{wang2024stablesr,lin2024diffbir,wu2024osediff,zhu2024oftsr} address this challenge by introducing powerful generative priors, achieving substantial progress in both restoration fidelity and perceptual quality.

Despite this progress, most existing methods are designed to produce a
single canonical restoration, implicitly assuming that the same detail
preference applies to different images and users. In practice, however,
the desired appearance is highly dependent on image content and user
intent. For scenes containing fine and irregular structures, such as
hair, animal fur, foliage, or weathered surfaces, users may prefer richer
texture expression and stronger detail enhancement. For portraits,
illustrations, or visually clean scenes, a smoother and more restrained
result may instead be desirable. Aggressive detail synthesis can introduce
distracting or semantically incorrect patterns, whereas overly conservative
restoration may suppress structures that the generative prior is capable
of recovering. Therefore, the goal is not simply to maximize the amount
of generated detail, but to adapt it to the preference of each restoration
scenario. This motivates an explicit and user-controllable detail dimension
rather than a single predetermined reconstruction.

Several recent methods have introduced controllability into generative super-resolution. StableSR~\cite{wang2024stablesr} provides a tunable quality-fidelity trade-off, ODTSR~\cite{fang2026odtsr} adjusts fidelity and prompt controllability through noise-conditioned visual streams, and IFCSR~\cite{back2026ifcsr} controls the balance between fidelity and realism through image-space manipulation. These advances demonstrate that users can benefit from multiple restoration preferences rather than a single predetermined output. Nevertheless, existing control variables often influence several image properties simultaneously, including degradation removal, texture strength, structural fidelity, generation stochasticity, and semantic response to prompts. Consequently, it can be difficult to interpret the control parameter specifically as a preference for the amount of image detail.

This motivates a complementary question to conventional real-world super-resolution: in addition to improving restoration quality, can a model allow users to continuously adjust the amount of expressed detail while preserving the same observed content, semantic instruction, and general restoration capability? Such a control should not require retraining a separate restoration model for every preference. More importantly, its effect should be distinguishable from simply changing the severity of the input degradation, the prompt, or the sampling randomness.

To this end, we introduce \textbf{DetailScale-SR}, a controllable
Real-ISR framework. Our key idea is to separate general restoration
capability from user-specified detail preference. We first establish a
robust SR pathway for recovering clean high-resolution content, and
subsequently learn continuous detail control without altering the
underlying restoration model. Given the same low-resolution image and
restoration instruction, users can adjust a scalar detail value to obtain
outputs ranging from restrained reconstruction to increasingly detailed
synthesis. We additionally provide an optional scale-aware guidance
mechanism that further amplifies the learned detail direction when stronger
enhancement is desired.

Overall, DetailScale-SR provides three complementary capabilities.
(1) It enables continuous detail control, allowing users to move from
restrained restoration to increasingly detailed synthesis and optional
enhancement according to their preferences.
(2) It places particular emphasis on text-sensitive restoration. Character
strokes, especially dense Chinese characters, are easily corrupted by blur,
noise, compression, and downsampling. Existing generative SR methods may
treat them as ordinary textures, producing distorted glyphs, substituted
characters, or visually plausible but semantically incorrect text~\cite{ma2022tatt,zhang2023difftsr,fang2026odtsr}. Through
text-bearing training examples and optional visible-text instructions,
DetailScale-SR combines degraded visual evidence with semantic cues to
improve both textual content recovery and spatial placement in posters,
storefronts, road signs, and other text-rich scenes.
(3) It further supports layer-oriented RGBA super-resolution, extending
restoration to transparent foreground content while preserving character
layout, foreground appearance, transparency boundaries, and alpha-layer
consistency. Together, these capabilities provide a flexible restoration
interface for natural images, text-rich scenes, and layered visual content.

Our contributions are summarized as follows:
\begin{itemize}
    \item We extend Real-ISR with a continuous detail-preference dimension, allowing users to adjust the amount of expressed detail while preserving the same input content and restoration model.

    \item We propose a two-stage controllable restoration framework that separates general super-resolution capability from detail-level modulation, together with an optional scale-aware guidance mechanism for stronger detail enhancement.

    \item We emphasize prompt-assisted text restoration and extend the evaluation beyond conventional RGB images to layer-oriented RGBA content, demonstrating the potential of generative SR for text-sensitive and transparency-aware restoration scenarios.
\end{itemize}

%% file: sec/2_relatedwork.tex
\section{Related Work}
\label{sec:related_work}

\subsection{Generative Image Super-Resolution}

Real-world image super-resolution aims to recover high-quality images under complex and unknown degradations~\cite{cai2019realsr,wei2020drealsr}.
BSRGAN~\cite{zhang2021bsrgan} and Real-ESRGAN~\cite{wang2021realesrgan} improve robustness through practical degradation synthesis.
Pretrained diffusion models further introduce strong generative priors for recovering plausible structures and textures~\cite{rombach2022ldm}.
StableSR~\cite{wang2024stablesr} introduces time-aware conditioning, while DiffBIR~\cite{lin2024diffbir} separates degradation removal from information regeneration.
PASD~\cite{yang2024pasd} and SeeSR~\cite{wu2024seesr} incorporate structural or semantic guidance, whereas SUPIR~\cite{yu2024supir} investigates model scaling and restoration-guided sampling.
DiT4SR~\cite{duan2025dit4sr} extends generative SR to diffusion transformers, while OSEDiff~\cite{wu2024osediff} and TSD-SR~\cite{dong2025tsdsr} improve efficiency through distillation-based restoration.

Most SR methods produce results at a fixed operating point, although the preferred amount of detail may vary across scenes and users.
Natural textures and animal fur may benefit from rich detail synthesis, while portraits and graphic designs may favor smoother restoration.
Deep Network Interpolation~\cite{wang2019dni} adjusts restoration behavior through parameter interpolation, while CFSNet~\cite{wang2019cfsnet} combines features associated with different objectives.
Recent diffusion-based methods provide further control over fidelity and generative enhancement.
StableSR~\cite{wang2024stablesr} and DiffBIR~\cite{lin2024diffbir} expose adjustable restoration guidance; PiSA-SR~\cite{sun2025pisasr} combines specialized LoRAs with tunable guidance; OFTSR~\cite{zhu2024oftsr} and RCOD~\cite{wu2025rcod} control the fidelity--realism trade-off in flow or latent spaces.
IFCSR~\cite{back2026ifcsr} combines fidelity- and realism-oriented predictions, while ODTSR~\cite{fang2026odtsr} introduces adjustable noise conditioning and fidelity-aware training.
Our method instead continuously regulates detail expression within a shared restoration pathway and supports stronger enhancement through optional scale-aware guidance.

Text restoration is another important SR application, particularly for posters, signs, and advertisements.
TATT~\cite{ma2022tatt} and DiffTSR~\cite{zhang2023difftsr} introduce text-specific priors or joint image--text modeling, while ODTSR~\cite{fang2026odtsr} demonstrates prompt-assisted Chinese text restoration.
Nevertheless, complex characters and dense layouts may still produce distorted strokes, substituted characters, or inaccurate spatial arrangements.
We preserve the text-rendering prior of the pretrained image-editing model~\cite{wu2025qwenimage,qwen2026image21} and further adapt it with text-rich restoration data, emphasizing the accuracy of textual content and spatial placement.

\subsection{Conditional Control in Generative Models}

Controllable generation commonly introduces additional conditions or lightweight components into pretrained generative models.
LoRA~\cite{hu2022lora} learns low-rank parameter updates for efficient task, concept, or style adaptation.
ControlNet~\cite{zhang2023controlnet} and T2I-Adapter~\cite{mou2023t2iadapter} inject structural conditions such as edges, depth, and pose, while IP-Adapter~\cite{ye2023ipadapter} incorporates reference-image features through an additional attention pathway.
GLIGEN~\cite{li2023gligen} introduces grounded spatial conditions, and Uni-ControlNet~\cite{zhao2023unicontrolnet} combines local and global controls within a shared framework.
Concept Sliders~\cite{gandikota2024conceptsliders} instead learn scalable parameter directions for continuous semantic control.

Other methods represent control signals through feature modulation or learnable context tokens.
Adaptive normalization~\cite{huang2017adain,perez2018film} converts conditions into layer-wise modulation parameters, whereas token-based methods~\cite{gal2023textualinversion} inject learned representations into attention layers.
Our Scale Template follows this lightweight conditioning principle by converting a scalar detail preference into layer-wise attention conditions.
Unlike spatial control maps or concept-specific adapters, it represents a continuous restoration preference and enables different levels of detail expression without changing the shared restoration model.

%% file: sec/3_method.tex
\section{Method}
\label{sec:method}

\subsection{Problem Formulation and Overview}
\label{sec:formulation}

Let $I\in\mathbb{R}^{H\times W\times C}$ denote a high-resolution image and $x=\mathcal{D}(I;\omega)$ its low-resolution observation, where $\mathcal{D}$ combines spatial reduction and real-world distortions, and $\omega$ denotes the sampled degradation parameters. Given $x$, an optional text instruction $p$, and a user-specified detail value $s\in[0,1]$, our goal is to reconstruct
\begin{equation}
    \hat I_s
    =
    \mathcal{R}_{\Theta,\Delta}
    \bigl(x,p;\Phi(s)\bigr),
    \label{eq:sr_task}
\end{equation}
where $\Theta$ denotes the pretrained image-editing model, $\Delta$ is an SR LoRA for image restoration, and $\Phi$ is a Scale Template that maps $s$ to an internal detail condition.

The desired variation in $\hat I_s$ concerns the amount of fine structure and texture, rather than the scene identity, spatial layout, or readable text. Unlike image-space interpolation between separately generated endpoint images, our method directly produces the result associated with the requested detail value. The restoration input remains identical across different values of $s$, requiring the model to learn different detail preferences for the same observed content.

\begin{figure*}[t]
    \centering
    \includegraphics[width=\textwidth]{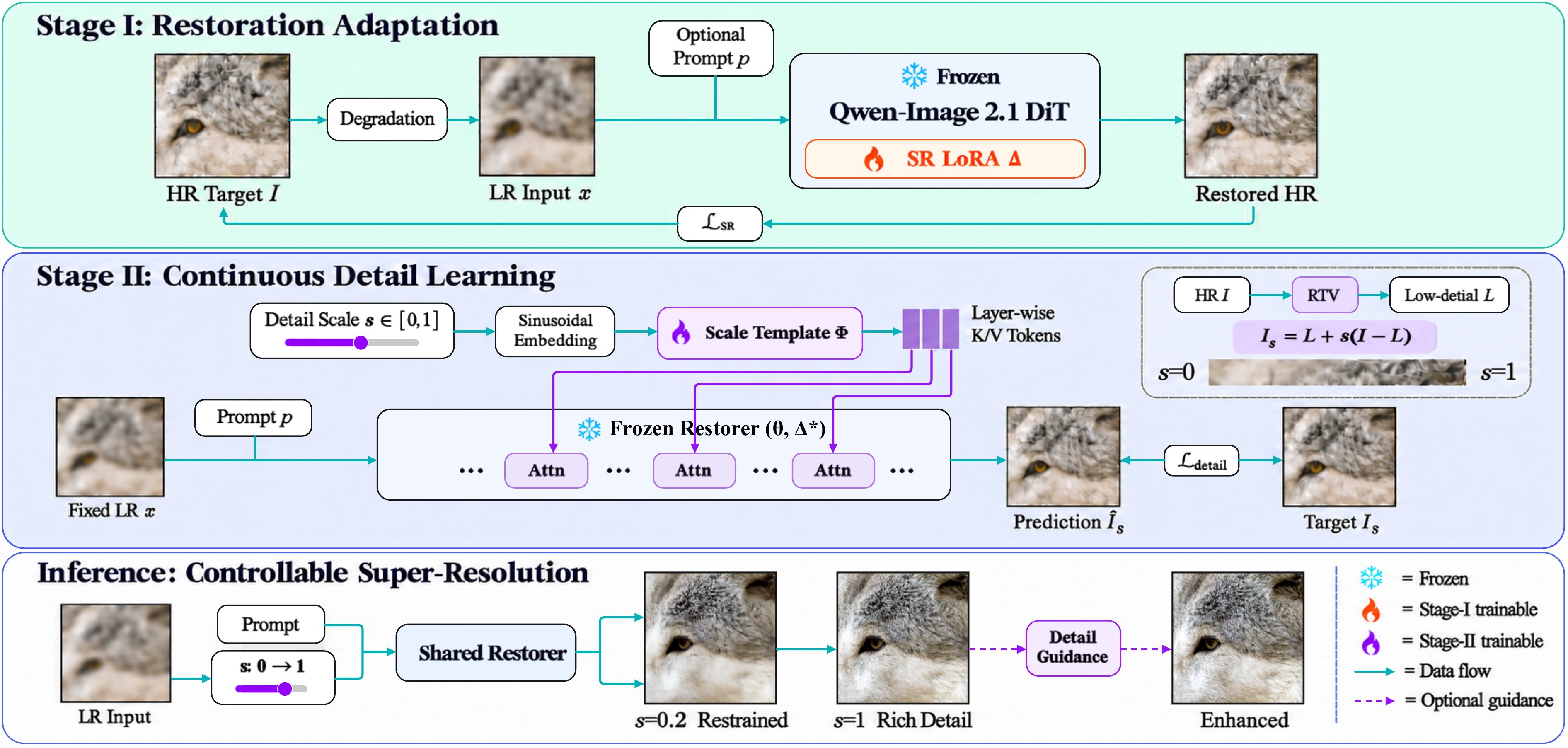}
    \caption{Overview of the proposed DetailScale-SR framework.}
    \label{fig:architecture}
\end{figure*}

As illustrated in Fig.~\ref{fig:architecture}, our framework follows a two-stage design. The first stage trains the SR LoRA $\Delta$ to adapt the pretrained model to real-world image restoration. The second stage freezes the established restoration capability and trains the Scale Template $\Phi$ for continuous detail control. This separation enables a single shared restorer to generate different levels of detail without training separate restoration models. The training objectives and conditioning mechanism are presented in the following sections.

\subsection{Stage I: Shared Restoration Adaptation}
\label{sec:stage1}

\paragraph{Restoration adaptation.}
We adopt Qwen-Image 2.1~\cite{qwen2026image21,wu2025qwenimage} as the image-editing backbone and introduce a rank-64 LoRA~\cite{hu2022lora} to establish a shared restoration capability. Following common practice in real-world SR~\cite{zhang2021bsrgan,wang2021realesrgan,fang2026odtsr}, each high-resolution target $I$ is transformed into a low-resolution observation $x$ through a combination of spatial downsampling and sampled real-world degradations. The resulting $\times4$ low-resolution image is bicubically resized to the target canvas and used as the spatially aligned restoration condition $u(x)$. This resizing aligns the condition with the output space without recovering the lost high-frequency information.

The editing instruction guides the model to remove degradation and recover plausible details while preserving the observed scene. Depending on the available user input, $p$ can be either a generic restoration instruction or an optional content-aware description. The latter can provide useful semantic cues in challenging cases, particularly for text-rich images, but is not required for standard restoration.

\paragraph{Content-anchored objective.}
We retain the standard conditional flow-matching formulation~\cite{lipman2023flowmatching} of the pretrained backbone and introduce no additional objective. The low-resolution observation and optional instruction are treated as conditions of velocity field, while the high-resolution image defines the clean trajectory endpoint. Let $z=\mathcal{E}(I)$ denote the latent representation of the target image, and let $z_t=\alpha_t z+\sigma_t\epsilon$ be the noisy latent at timestep $t$, where $\epsilon\sim\mathcal{N}(0,\mathbf I)$. We freeze the pretrained parameters $\theta$ and optimize only the restoration adapter $\Delta$ using
\begin{equation}
\mathcal{L}_{\mathrm{SR}}(\Delta)=
\mathbb{E}_{I,\omega,p,t,\epsilon}
\left\|
v_{\theta,\Delta}(z_t,t\mid u(x),p)-v_t^\star
\right\|_2^2 ,
\label{eq:stage1}
\end{equation}
where $v_t^\star$ denotes the target trajectory velocity. The noise schedule and its coefficients $\alpha_t$ and $\sigma_t$ follow the original Qwen-Image training formulation. Because the objective is evaluated across sampled timesteps, the adapter learns to redirect the complete conditional trajectory from degraded observations toward clean image latents rather than matching only the final decoded output. Freezing $\theta$ confines this adaptation to the low-rank restoration pathway and preserves the generative and semantic priors of the pretrained model. Stage~I establishes the shared restoration function used across all detail settings. During Stage~II, both $\theta$ and $\Delta$ remain fixed, ensuring that variations induced by $s$ are learned through the detail-conditioning module rather than changes to the underlying restorer.

\subsection{Stage II: Continuous Detail Conditioning}
\label{sec:stage2}

\paragraph{Continuous detail targets.}
To assign a consistent meaning to the control value $s$, we construct multiple aligned targets for the same low-resolution observation. Given a high-resolution image $I$, we first obtain a low-detail endpoint
\begin{equation}
    L=\mathcal{T}_{\mathrm{RTV}}(I),
\end{equation}
where $\mathcal{T}_{\mathrm{RTV}}$ denotes relative total variation (RTV) filtering~\cite{xu2012rtv}. RTV suppresses small-scale textures while largely preserving prominent structures, providing a practical reference for reduced detail expression. It is used only to construct the supervision target and does not alter the low-resolution observation.

For a sampled detail value $s\in[0,1]$, we define the corresponding target as
\begin{equation}
    I_s=(1-s)L+sI=L+s(I-L).
    \label{eq:detail_target}
\end{equation}
The two endpoints represent texture-suppressed and full-detail targets, while intermediate values form a spatially aligned and continuous supervision path. The low-resolution condition is always generated from the original image as $x=\mathcal{D}(I;\omega)$, independently of $s$. Therefore, the model must learn the association between $s$ and the desired detail expression instead of inferring it from changes in the input degradation. Note that interpolation is used only to construct training targets; at inference time, the output for a given $s$ is generated directly by the model rather than blended from endpoint predictions.

\paragraph{Detail Controlling Module.}
We introduce a trainable Detail Controlling Module $\Phi$ to translate the scalar $s$ into internal attention conditions. The scalar is first encoded using a sinusoidal embedding~\cite{vaswani2017attention},
\begin{equation}
    e(s)=\operatorname{PE}(s),
\end{equation}
and then mapped to block-specific key and value tokens:
\begin{equation}
\begin{aligned}
    K_b^{\mathrm{tmp}}(s)
    &=
    \operatorname{reshape}
    \left(f_b^K(e(s))+P_b^K\right),\\
    V_b^{\mathrm{tmp}}(s)
    &=
    \operatorname{reshape}
    \left(f_b^V(e(s))+P_b^V\right),
\end{aligned}
\label{eq:template}
\end{equation}
where $f_b^K$ and $f_b^V$ are learnable encoders for transformer block $b$, and $P_b^K$ and $P_b^V$ are learned token-position embeddings. Since these mappings operate continuously on $s$, the conditioning is not restricted to a predefined set of discrete detail categories. 
The generated Template tokens are appended to the original attention keys and values:
\begin{equation}
\begin{aligned}
    \widetilde K_b(s)
    &=[K_b^{\mathrm{orig}};K_b^{\mathrm{tmp}}(s)],\\
    \widetilde V_b(s)
    &=[V_b^{\mathrm{orig}};V_b^{\mathrm{tmp}}(s)],\\
    \operatorname{Attn}_b(s)
    &=
    \operatorname{softmax}
    \left(
    \frac{Q_b\widetilde K_b(s)^\top}{\sqrt d}
    +\widetilde M_b
    \right)
    \widetilde V_b(s),
\end{aligned}
\label{eq:kv_injection}
\end{equation}
where $\widetilde M_b$ is the attention mask extended for the additional tokens. The original image, text, and editing conditions are retained; the Template provides an additional pathway through which the requested detail preference can influence feature synthesis.

\paragraph{Training objective.}
Stage~II follows the same flow-matching objective as Stage~I, with the original target $I$ replaced by the scale-conditioned target $I_s$. The low-resolution condition $u(x)$, the instruction $p$, and the backbone noise schedule remain identical to Stage~I; the only additional conditioning signal is the scale-dependent representation $\Phi(s)$. Specifically, we encode $I_s$ as $z_s=\mathcal{E}(I_s)$ and construct its noisy latent using the original backbone schedule. The pretrained parameters $\theta$ and the Stage~I restoration adapter $\Delta^\star$ remain frozen, while only the Scale Template $\Phi$ is optimized through the conditioning $\Phi(s)$. Since $x$ is generated independently of $s$, the Template cannot infer the requested detail level from the degradation strength. Instead, it must use the scale-conditioned attention tokens to account for variations among the aligned targets. Sampling $s$ continuously within $[0,1]$ exposes both endpoint and intermediate preferences to the same shared restorer, encouraging a continuous rather than categorical representation of detail. Therefore, the two stages differ in their supervision targets and trainable parameters, rather than in the underlying flow-matching formulation.

\subsection{Scale-Aware Detail Guidance}
\label{sec:guidance}

The continuous detail condition enables direct inference at any requested value $s\in[0,1]$. At each sampling step, the scale-conditioned velocity prediction is written as
\begin{equation}
    v(s)=v_{\theta,\Delta^{\star},\Phi}
    \bigl(z_t,t\mid u(x),p;\Phi(s)\bigr).
    \label{eq:direct_scale}
\end{equation}
This direct mode provides continuous control using a single requested detail condition, without training separate restoration models for different values of $s$.

To further amplify the detail-enhancement capability learned by the Scale Template, we introduce an optional scale-aware detail guidance mechanism. Given a target detail value $s_+$ and a lower reference value $s_{\mathrm{ref}}<s_+$, we evaluate two branches using the same latent state, low-resolution observation, and text instruction. Their difference defines a detail-sensitive direction in the velocity field:
\begin{equation}
\begin{aligned}
    d_t^{\mathrm{detail}}
    &=v(s_+)-v(s_{\mathrm{ref}}),\\
    v_{\mathrm{guide}}
    &=v(s_+)+(\gamma-1)d_t^{\mathrm{detail}},
\end{aligned}
\label{eq:detail_guidance}
\end{equation}
where $\gamma\geq1$ controls the guidance strength. When $\gamma=1$, the formulation reduces to the direct prediction at $s_+$. Increasing $\gamma$ extrapolates along the learned detail-sensitive direction, encouraging stronger fine structures and richer textures than the direct branch alone.

This formulation is inspired by classifier-free guidance~\cite{ho2022cfg}, but the two branches differ in their detail conditions rather than their text prompts. Keeping $x$, $p$, $z_t$, and $t$ identical isolates the variation induced by the detail scale and avoids introducing an additional semantic direction. Moreover, the guidance strength $\gamma$ is separated from the detail value $s$: the Scale Template remains evaluated within its trained interval, while $\gamma$ controls how strongly the learned detail direction is expressed.

%% file: sec/4_Experiments.tex
\section{Experiments}
\label{sec:experiments}

\begin{table*}[!t]
\centering
\caption{Quantitative comparison on RealSR (top) and DRealSR (bottom). Best and
second-best values of each column are highlighted in bold and underlined.
\emph{Rank} is the average per-metric rank over PSNR, SSIM, LPIPS, DISTS and FID
(1 = best, lower is better); the no-reference metrics are reported but not ranked,
as they measure perceived quality rather than agreement with the reference.}
\label{tab:main}
\small
\setlength{\tabcolsep}{4pt}
\begin{tabular}{@{}llcccccccc@{}}
\toprule
Dataset & Method & PSNR$\uparrow$ & SSIM$\uparrow$ & LPIPS$\downarrow$ & DISTS$\downarrow$ & FID$\downarrow$ & MUSIQ$\uparrow$ & MANIQA$\uparrow$ & Rank$\downarrow$ \\
\midrule
\multirow{12}{*}{RealSR} & OSEDiff & 25.15 & 0.7341 & 0.2921 & 0.2128 & 123.49 & 69.09 & 0.6326 & 7.70 \\
 & PiSA-SR & 25.50 & 0.7417 & 0.2672 & 0.2044 & 124.09 & \underline{70.15} & 0.6560 & 5.80 \\
 & TSD-SR & 24.81 & 0.7172 & 0.2743 & 0.2104 & 114.45 & \textbf{71.19} & 0.6347 & 8.00 \\
 & TVT & 25.81 & 0.7596 & 0.2597 & 0.2061 & 109.90 & 69.89 & 0.6232 & 4.40 \\
 & SUPIR & 23.65 & 0.6620 & 0.3541 & 0.2488 & 130.38 & 62.09 & 0.5780 & 11.80 \\
 & DiT4SR & 23.50 & 0.6657 & 0.3215 & 0.2251 & 118.55 & 67.76 & \underline{0.6564} & 10.60 \\
 & SeeSR & 25.15 & 0.7210 & 0.3007 & 0.2225 & 125.63 & 69.82 & 0.6444 & 9.10 \\
 & ODTSR & 25.07 & 0.7361 & 0.2398 & 0.1894 & 101.49 & 68.29 & \textbf{0.6622} & 4.60 \\
 & IFCSR ($\gamma{=}0.5$) & 25.14 & 0.7055 & 0.2866 & 0.2141 & 110.60 & 69.68 & 0.6293 & 8.20 \\
 & Ours, $s{=}0.2$ & \textbf{26.52} & \textbf{0.7725} & 0.2433 & 0.2029 & 109.58 & 61.49 & 0.5805 & \underline{2.80} \\
 & Ours, $s{=}0.5$ & \underline{26.25} & \underline{0.7610} & \textbf{0.2267} & \underline{0.1837} & \underline{94.40} & 63.32 & 0.6123 & \textbf{1.80} \\
 & Ours, $s{=}1.0$ & 25.48 & 0.7300 & \underline{0.2343} & \textbf{0.1834} & \textbf{88.40} & 65.53 & 0.6368 & 3.20 \\
\midrule
\multirow{12}{*}{DRealSR} & OSEDiff & 27.92 & 0.7835 & 0.2968 & 0.2165 & 135.29 & 64.65 & 0.5895 & 6.40 \\
 & PiSA-SR & 28.32 & 0.7804 & 0.2960 & 0.2169 & 130.48 & 66.11 & 0.6146 & 5.00 \\
 & TSD-SR & 27.77 & 0.7559 & 0.2966 & 0.2136 & 135.32 & \textbf{66.61} & 0.5850 & 7.20 \\
 & TVT & 28.27 & 0.7899 & 0.2899 & 0.2205 & 134.28 & 65.56 & 0.5776 & 5.20 \\
 & SUPIR & 25.09 & 0.6460 & 0.4243 & 0.2795 & 169.48 & 58.79 & 0.5471 & 12.00 \\
 & DiT4SR & 25.40 & 0.6657 & 0.3869 & 0.2508 & 163.05 & 65.75 & \textbf{0.6287} & 11.00 \\
 & SeeSR & 28.07 & 0.7683 & 0.3174 & 0.2315 & 147.39 & 65.09 & 0.6044 & 8.20 \\
 & ODTSR & 28.14 & 0.7736 & \textbf{0.2592} & \textbf{0.1926} & \textbf{119.86} & 62.86 & \underline{0.6227} & \underline{2.80} \\
 & IFCSR ($\gamma{=}0.5$) & 27.60 & 0.7371 & 0.3278 & 0.2361 & 133.84 & \underline{66.57} & 0.5940 & 9.00 \\
 & Ours, $s{=}0.2$ & \textbf{28.91} & \textbf{0.8144} & 0.2741 & 0.2170 & 129.66 & 58.63 & 0.5638 & 3.20 \\
 & Ours, $s{=}0.5$ & \underline{28.55} & \underline{0.7943} & \underline{0.2686} & \underline{0.2030} & 122.54 & 60.67 & 0.5909 & \textbf{2.20} \\
 & Ours, $s{=}1.0$ & 27.50 & 0.7513 & 0.2908 & 0.2058 & \underline{120.99} & 63.35 & 0.6130 & 5.80 \\
\bottomrule
\end{tabular}
\end{table*}

\subsection{Experimental Settings}
\label{sec:experimental_settings}

\paragraph{Implementation.}
We build our model upon Qwen-Image 2.1~\cite{qwen2026image21,wu2025qwenimage}.
In Stage~I, a rank-64 LoRA~\cite{hu2022lora} is trained for real-world image
restoration, while Stage~II freezes the backbone and SR LoRA and optimizes only
the Scale Template. Unlike patch-based training, we use complete images with
their aspect ratios preserved, resizing them only to satisfy the resolution
budget. We adopt AdamW~\cite{loshchilov2019adamw} with a learning rate of
$10^{-4}$, and train both stages on four accelerators. At inference, we use the default sampling configuration with $s{=}1$. Other details are provided in the
supplementary material.

\paragraph{Datasets.}
We focus on the $\times4$ real-world image super-resolution task. We train Stage~I on 40,000 LSDIR~\cite{li2023lsdir} images and 2,000 text-rich samples. Unlike previous methods that train on cropped patches, we use complete images with their aspect ratios preserved. LQ inputs are synthesized using blur, downsampling, noise, and compression. In Stage~II, RTV~\cite{xu2012rtv} is applied to construct continuous detail targets. We evaluate our method on the RealSR~\cite{cai2019realsr} and DRealSR~\cite{wei2020drealsr} benchmarks.

\paragraph{Metrics.}
Following previous real-world image super-resolution works, we report PSNR and SSIM~\cite{wang2004ssim} on the luminance channel to measure reconstruction fidelity. LPIPS~\cite{zhang2018lpips} and DISTS~\cite{ding2022dists} evaluate full-reference perceptual similarity, while FID~\cite{heusel2017fid} measures the distributional discrepancy between restored images and ground-truth images. We additionally report MUSIQ~\cite{ke2021musiq} and MANIQA~\cite{yang2022maniqa} as no-reference measures of perceptual image quality.

\paragraph{Compared Methods.}
We compare our method with representative real-world image super-resolution approaches, including OSEDiff~\cite{wu2024osediff}, PiSA-SR~\cite{sun2025pisasr}, TSD-SR~\cite{dong2025tsdsr}, TVT~\cite{yi2025tvt}, SUPIR~\cite{yu2024supir}, DiT4SR~\cite{duan2025dit4sr}, and ODTSR~\cite{fang2026odtsr}. For each method, we use its official implementation, released checkpoint, and recommended inference settings.

\subsection{Comparison with State-of-the-Art Methods}
\label{sec:exp_comparison}

\paragraph{Quantitative Comparisons.}
Table~\ref{tab:main} compares nine representative Real-ISR methods with
our model at three detail levels on RealSR and DRealSR. All our results
are produced by a single checkpoint with the backbone and restoration
adapter frozen; only the detail level $s$ and guidance strength $\gamma$
are varied. The control variable is therefore isolated from restoration
capability, and the table evaluates one model over a continuous range
rather than multiple separately configured systems. Our method achieves
the best or second-best LPIPS, DISTS, and FID on both datasets, while its
most conservative setting obtains the highest PSNR and SSIM. To summarize
the reference-based performance, we average the per-metric ranks over
PSNR, SSIM, LPIPS, DISTS, and FID, under which our method ranks first on
both benchmarks. No-reference metrics are reported but excluded from this
ranking because they measure perceived quality rather than agreement with
the reference. These results show that a single scalar-conditioned
restorer can provide continuous detail control without sacrificing
endpoint fidelity, match specialized baselines across the
fidelity--detail spectrum, and remain adjustable at inference time.

\begin{figure*}[t]
\centering
\includegraphics[width=\textwidth]{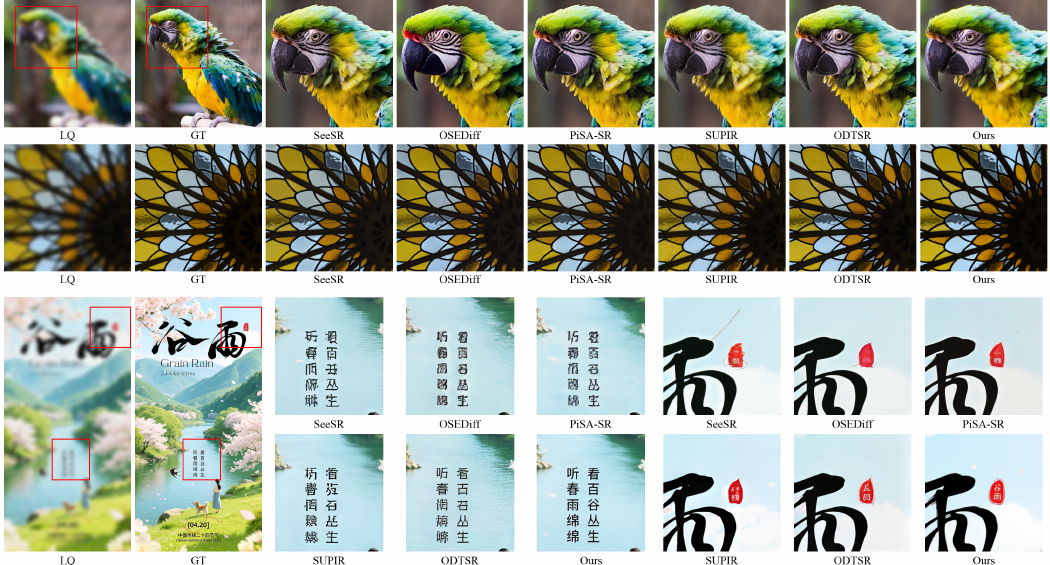}
\caption{Qualitative comparison on $\times4$ real-world super-resolution.Our method performs favorably overall and is especially strong on text-rich content.}
\label{fig:qualitative}
\end{figure*}

\paragraph{Qualitative Comparisons.}
\Cref{fig:qualitative} shows qualitative comparisons on natural
texture, thin structures and text-rich content, all produced from the
same low-resolution inputs. The first two rows show that our method
produces high-quality restoration in fine-grained textures and thin
high-contrast structures under severe degradation: fine detail is
recovered without obvious artifacts, edges that should be
straight remain straight and continuous, and the restored colours
stay close to those of the reference. Besides, the third row
demonstrates our capability on text-rich content, especially in
Chinese scenes, where characters are restored with correct glyph
shapes and connected strokes, and where small print and small
graphic elements remain readable. The same behaviour is visible in
the enlarged text and seal regions, which contain dense strokes and
fine boundaries that are easily lost during degradation. Moreover,
the quality can be further boosted with a text annotation prompt,
indicating the prompt controllability of our framework.
% 需要：\usepackage{booktabs} \usepackage{multirow} \usepackage[table]{xcolor}

\subsection{Continuous Detail Control and Enhancement}
\label{sec:exp_control}

\begin{figure}[t]
  \centering
  \includegraphics[width=\columnwidth]{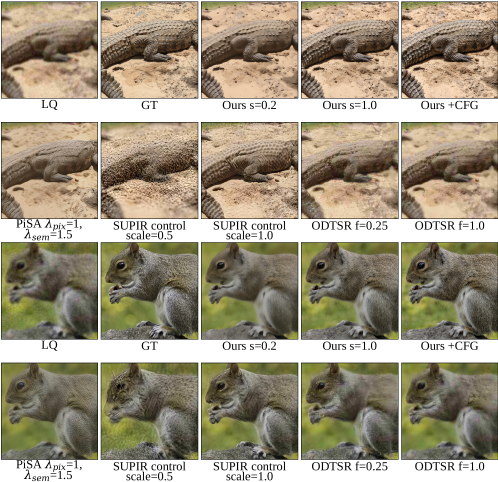}
\caption{Controllable SR results on two heavily degraded images; Our
method delivers clearer detail-controllable generation and preserves structure
better under heavy degradation.}
  \label{fig:qual_control}
\end{figure}

\begin{table}[!t]
\centering
\caption{Continuous detail control on RealSR (top) and DRealSR (bottom). The last
row of each block applies Detail-CFG.}
\label{tab:control}
\scriptsize
\setlength{\tabcolsep}{2pt}
\renewcommand{\arraystretch}{0.95}
\resizebox{\columnwidth}{!}{%
\begin{tabular}{@{}lcccccc@{}}
\toprule
Setting & PSNR$\uparrow$ & SSIM$\uparrow$ & LPIPS$\downarrow$ & MANIQA$\uparrow$ & MUSIQ$\uparrow$ & FID$\downarrow$ \\
\midrule
$s{=}0.0$ & \textbf{26.69} & \textbf{0.7723} & 0.2766 & 0.5533 & 60.94 & 130.46 \\[1pt]
$s{=}0.4$ & 26.56 & 0.7685 & 0.2292 & 0.6030 & 63.00 & 98.30 \\[1pt]
$s{=}0.8$ & 26.14 & 0.7514 & \textbf{0.2215} & 0.6251 & 64.26 & \textbf{87.44} \\[1pt]
$s{=}1.0$ & 25.48 & 0.7300 & 0.2343 & 0.6368 & 65.53 & 88.40 \\[1pt]
CFG $\gamma{=}2$ & 24.06 & 0.6587 & 0.2835 & \textbf{0.6653} & \textbf{67.86} & 98.53 \\[1pt]
\midrule
$s{=}0.0$ & \textbf{29.08} & \textbf{0.8243} & 0.2985 & 0.5346 & 57.01 & 147.76 \\[1pt]
$s{=}0.4$ & 28.78 & 0.8021 & \textbf{0.2715} & 0.5840 & 60.33 & 125.92 \\[1pt]
$s{=}0.8$ & 28.09 & 0.7736 & 0.2783 & 0.6022 & 61.65 & 121.18 \\[1pt]
$s{=}1.0$ & 27.50 & 0.7513 & 0.2908 & 0.6130 & 63.35 & \textbf{120.99} \\[1pt]
CFG $\gamma{=}2$ & 25.56 & 0.6606 & 0.3608 & \textbf{0.6314} & \textbf{66.05} & 138.63 \\[1pt]
\bottomrule
\end{tabular}%
}
\end{table}

To observe the controllability more clearly and to verify the robustness,
we conduct the controllable SR experiment under a stronger degradation
setting (Fig.~\ref{fig:qual_control}). As the detail level $s$ increases,
the ridges on crocodile armour and strands of the squirrel's fur
are progressively sharpened, while pose, geometry and colour remain
unchanged. The compared methods exhibit two main limitations: their
results tend to be over-smoothed with inconspicuous fine detail, and their
robustness deteriorates markedly under severe degradation. When the CFG
mechanism is enabled, the detail condition is extrapolated along the
learned direction, so that the produced texture becomes richer than the
ground truth and the model performs \emph{enhancement} beyond restoration.

Table~\ref{tab:control} quantifies this behaviour on both benchmarks.
Sweeping $s$ from 0 to 1 traces a continuous path: PSNR and SSIM decrease
monotonically while MUSIQ and MANIQA increase, and FID reaches its minimum
at a large detail level, so the model moves smoothly from
fidelity-oriented restoration to detail-oriented synthesis. LPIPS attains
its best value at an intermediate scale ($s{=}0.8$ on RealSR and $s{=}0.4$
on DRealSR), indicating that moderate detail generation keeps the closest
feature-space correspondence. Detail-CFG (last row of each block) further
improves all no-reference quality metrics, extending the controllable range
beyond standard restoration and enabling \emph{enhancement} with
perceptually richer details than the ground-truth reference.
\subsection{Prompt-Assisted Text Restoration}
\label{sec:exp_text}
\begin{figure}[t]
  \centering
  \includegraphics[width=\columnwidth]{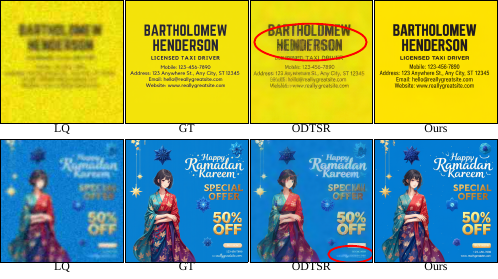}
  \caption{Text restoration with prompt assistance. Our method performs
  favorably on challenging cases containing many texts and recovers the
  content accurately; more examples are provided in the supplementary
  material.}
  \label{fig:text}
\end{figure}
Our method also performs strongly on text rendering, especially for images
that contain a large amount of text and require accurate spatial
correspondence. Under heavy degradation, some characters cannot be
recovered precisely from the low-resolution observation alone, so we adopt
a prompt-assisted strategy that supplies the text content as a prompt. In
this setting our method recovers both the character content and its spatial
position accurately, whereas the compared methods tend to produce blurred
characters or spatially mismatched text (Fig.~\ref{fig:text}). In addition,
benefiting from the four-channel RGBA generation capability of
Qwen-Image~2.1~\cite{qwen2026image21,wu2025qwenimage}, our approach can perform layer-wise super-resolution or
editing as needed, for example in poster-style scenes.

\subsection{Ablation on Two-Stage Training Strategy}
\label{sec:ablation_twostage}

Our training strategy freezes the Stage-I super resolution LoRA and optimizes only the
detail template in Stage II. To justify this two-stage design, we train a joint
variant that optimizes both components together, starting from the same Stage-I
initialization and using identical data, targets and budget, so that the freeze
is the only difference between the two models.
\begin{table}[t]
\centering
\caption{Ablation on the two-stage training strategy on RealSR }
\label{tab:ablation_twostage}
\small
\setlength{\tabcolsep}{3pt}   
\begin{tabular}{@{}lccccc@{}}
\toprule
Strategy & PSNR$\uparrow$ & SSIM$\uparrow$ & LPIPS$\downarrow$ & DISTS$\downarrow$ & FID$\downarrow$ \\
\midrule
Two-stage (frozen) & \textbf{25.48} & \textbf{0.7300} & \textbf{0.2343} & \textbf{0.1834} & \textbf{88.40} \\
Joint (one-stage)  & 23.50 & 0.6605 & 0.2972 & 0.2171 & 102.77 \\
\bottomrule
\end{tabular}
\end{table}
As shown in Table~\ref{tab:ablation_twostage}, joint optimization degrades all
fidelity metrics while yielding only slightly higher no-reference scores: the
detail objective then rewrites the restoration capability instead of modulating
it. Freezing the Stage-I LoRA confines the Stage-II update to the conditioning
pathway, which keeps the restorer intact across detail levels.

\label{sec:exp_ablation}

%% file: sec/5_conclusion.tex
\section{Conclusion}
\label{sec:conclusion}
We propose DetailScale-SR, a controllable diffusion framework for
real-world image super-resolution that separates restoration
capability from detail expression. Unlike existing controllable
methods that entangle detail with prompts, sampling noise or
degradation severity, we inject a continuous scalar into the
attention layers of a frozen restoration model, so that the amount of
expressed detail can be adjusted without any additional inference.
Built upon this formulation, the two-stage training scheme yields a
continuous and monotonic detail spectrum within a single model.
Extensive experiments on RealSR and DRealSR show that DetailScale-SR
is competitive in fidelity, while
preserving typography in text-rich scenes.